\documentclass{article}
\usepackage[utf8]{inputenc}
\usepackage{keyval}
\usepackage{nips11,times}
\usepackage{amsthm}
\usepackage[compress]{natbib}
\usepackage{amsmath}
\usepackage{amssymb}
\usepackage{xspace}
\usepackage{array}
\usepackage{graphicx}
\usepackage{wrapfig}

\newcounter{lemmac}
\newcounter{defnc}

\newtheorem{theorem}{Theorem}

\newtheorem{prop}[lemmac]{Proposition}

\newtheorem{prob}[lemmac]{Problem}
\theoremstyle{definition}
\newtheorem{defn}[defnc]{Definition}

\newcommand{\Gc}{\mathcal{G}}
\newcommand{\reals}{\mathbb{R}}
\DeclareMathOperator{\st}{ST}
\DeclareMathOperator{\Avcost}{Avcost}

\title{Statistical estimation for optimization problems on graphs}

\author{
Mikhail Langovoy and Suvrit Sra\\
Max Planck Institute for Intelligent Systems\\
72076 T\"ubingen, Germany\\
\texttt{\{langovoy, suvrit.sra\}@tuebingen.mpg.de}\\
}

\nipsfinalcopy 

\begin{document}
\maketitle
\vspace*{-10pt}
\begin{abstract}
  Large graphs abound in machine learning, data mining, and several related
  areas. A useful step towards analyzing such graphs is that of obtaining
  certain summary statistics---e.g., or the expected length of a shortest path
  between two nodes, or the expected weight of a minimum spanning tree of the
  graph, etc. These statistics provide insight into the structure of a graph,
  and they can help predict global properties of a graph. Motivated thus, we
  propose to study statistical properties of structured subgraphs (of a given
  graph), in particular, to estimate the expected objective function value of
  a combinatorial optimization problem over these subgraphs. The general task
  is very difficult, if not unsolvable; so for concreteness we describe a more
  specific statistical estimation problem based on spanning trees. We hope
  that our position paper encourages others to also study other types of
  graphical structures for which one can prove nontrivial statistical
  estimates.
\end{abstract}

\section{Introduction}
A cornucopia of applications in machine learning and related areas involve
large-scale graphs. Towards analyzing such graphs a basic step is that of
obtain certain summary statistics. For example, we might want to know what
might be the expected length of a shortest path between two nodes, or what is
the expected weight of an associated minimum spanning tree, etc. Such
statistics provide insight into the global structure of a graph; and
estimating them helps predict properties of the entire graph without having to
actually look at the whole graph, a very practical scenario.

Our considerations stem from a classic paper of~\citet{fri85}, who studied the
expected value of the weight of a minimum spanning tree (MST) of a complete
graph on $n$ nodes, with edge-weights distributed according to a common
distribution function. For such graphs, Frieze obtained an explicit value for
the expected weight of an MST as $n$ tends to infinity. In subsequent years,
his analysis has been refined and extended to cover more general graphs, and
under different assumptions---see~\citep{st02} and the references therein.

This precedent suggests that with increasing sizes, one can estimate
statistical properties of various combinatorial structures on graphs. This
statement brings us to the key challenge of this paper.
\begin{prob}[\bf Statistics on graphical structures]
  \label{prob.main}
  Let $G_n=(V_n,E)$ be a graph with $n$ vertices, and let $\mathcal{G}$ be a
  collection of certain ``structured'' subgraphs of $G_n$. Let $\varphi: \Gc
  \to \reals_+$ be function that measures the ``cost'' of a subgraph in
  $\Gc$. As $n$ tends to infinity, what can we say about the expected value
  $\mathbb{E}[\min_{g \in \Gc}\varphi(g)]$ of the minimum cost structure, and under
  what restrictions on the structures $\Gc$ and on the cost function
  $\varphi$?
\end{prob}

Our current paper is a position paper that advances Problem~\ref{prob.main} as
a key research question worthy of careful investigation. Admittedly, in
general this problem is very difficult; perhaps too broad to be useful. But
our ultimate aim is less to tackle the general problem and more to identify
special classes of \emph{structures} and \emph{cost functions}, for which we
can make nontrivial statistical statements. We hope that this workshop paper
stimulates discussion and also encourages others to study this problem.

\section{Formulation}
\label{sec.formulation}
Let us now move onto a somewhat more formal treatment of Problem~1.

Let $G_n = ( V_n , E_n )$ be a graph with $n$ vertices and $|E_n|$ edges. For any vertex $v \in V_n$ we observe an integer-valued random variable $X_v : \Omega \rightarrow \{ 1, 2, \ldots, k\}$ (on an appropriate probability space $\Omega$). Here $k$ is assumed to be finite, but otherwise unknown. We call the collection $\{ X_v \,|\, v \in V_n \}$ a \emph{random coloring} of the graph $G_n$.\footnote{We can also consider infinite colorings based on $X_v : \Omega \rightarrow \mathbb{N}$; but for simplicity we study only finite ones.}

Assume that $\{ X_v \,|\, v \in V_n \}$ form a collection of completely independent random variables and that they are identically distributed according to a distribution function $F_X$.

Further, suppose that for any pair of vertices $ v_i, v_j \in V_n$ we observe a real-valued random variable $Y_{v_i, v_j} : \quad \Omega \rightarrow \mathbb{R}_{+} $, where $\{ Y_{v_i, v_j} \,|\, v_i, v_j  \in V_n \}$ form a collection of completely independent random variables that are identically distributed according to a distribution function $F$. It is also assumed for now that the $X$'s and $Y$'s are completely independent of each other. We will add extra assumptions on $F$ and $F_X$ below, when necessary.

\textbf{Remark:} The above assumption on edge weights corresponds to the case when $G_n$ is a complete graph on $n$ vertices. To study general graphs on $n$ vertices, one has to consider only the reduced collection of random variables $\{ Y_{v_i, v_j} \,|\, (v_i, v_j)  \in E_n \}$.

Denote by $\mathcal{G} = \{ G \,|\, G \subseteq G_n \}$, a collection of (structured) subgraphs of $G_n$. Fix a \emph{cost function} $\varphi: \, \mathcal{G} \rightarrow \mathbb{R}_{+}$. Ultimately, we will be interested in the case when $\varphi$ is a set function (over the sets of vertices or edges participating in the subgraphs characterized by $\Gc$).

Often, a more convenient and specific form of $\varphi$ might be assumed, namely that for any $G = ( V , E ) \subseteq G_n$ there exists a decomposition

\begin{equation}
  \label{eq.1}
  \varphi (G) = \varphi ( \{ ( X_v, Y_e ) \,|\, v \in V, e \in E \} ) = \varphi_1 (\{ Y_e  \,|\, e \in E \}) + \varphi_2 (\{ X_v \,|\, v \in V \}),
\end{equation}
where $\varphi_1$ is a cost function that depends only on the edges, and $\varphi_2$ depends only on the vertices.

For concreteness, we now focus on the following statistical goal: \emph{estimate the average cost} (per vertex) of a minimal cost spanning tree of $G_n$---henceforth, $\varphi$-MST.\footnote{A similar discussion also applies to other problems such as shortest paths, cuts, etc.} We want a computationally efficient procedure for this estimate, and the estimator itself must be consistent. Additionally, we also care about the corresponding rates of convergence.

Since the original graph $G$ may be very large and a direct computation of the (minimal cost) spanning tree may be infeasible, we suggest computing an estimator based on a suitably constructed auxiliary graph that is much smaller than $G$, but exhibits similar statistical properties. To this end, we propose the following generic method.

\paragraph{Method}
\begin{enumerate}
\item On the basis of the coloring $\{ X_v \,|\, v \in V \}$, construct a suitable (possibly problem-dependent) $\sqrt{\, n \,}$-consistent estimate $\widehat{F}_X$ of the distribution function $F_X$.
\item Using the collection $\{ Y_e \,|\, e \in E \}$, construct a suitable $\sqrt{\, |E|    \,}$-consistent estimate $\widehat{F}$ of the edge weight distribution
  function $F$. Note that when $G_n$ is a complete graph, already the standard
  empirical distribution function gives an $n$-consistent estimate.
\item Generate an auxiliary graph $G_{d(n)}^{'} = ( V_{d(n)}^{'} ,  E_{d(n)}^{'} )$, having $d(n)$ vertices, where $d(n)$ is suitably chosen and satisfies the growth conditions
  \begin{equation}
    \label{eq.2}
    \lim_{n \rightarrow \infty} d(n) = \infty \,, \quad\quad \lim_{n \rightarrow \infty} \frac{\, d(n) \,}{\, n \,} = 0 \,.
  \end{equation}
\item Simulate i.i.d.\ random variables $\{ X_v^{'} \,|\, v^{'} \in
  V_{d(n)}^{'} \}$ and $\{ Y_e^{'} \,|\, e \in E_{d(n)}^{'} \}$ from the distribution function estimates constructed at Steps 1 and 2 correspondingly. (Remark: for
  a complete graph $G_n$, we generate $G_{d(n)}^{'}$ to be complete as well.)
\item Find the minimum $\varphi$-cost spanning tree  $\st (G_{d(n)}^{'})$; compute $\varphi (\st (G_{d(n)}^{'}))$. (Remark: This step requires solution of a potentially hard discrete optimization problem.)
\end{enumerate}

Based on the above generic method, we introduce the estimate:

\begin{equation}
  \label{eq.3}
\widehat{\Avcost} (G_n) \,:=\, \frac{\, \varphi (\st (G_{d(n)}^{'})) \,}{\, d(n) \,} \,.
\end{equation}

Processing the reduced graph $G_{d(n)}^{'}$ is obviously much faster than processing $G_n$ itself. But we need to theoretically characterize to what extent it is acceptable to process $G_{d(n)}^{'}$. To that end, we attempt to investigate the following main questions:
\begin{itemize}
\item When is the above method consistent;
\item What can be the rate of convergence of the estimator~\eqref{eq.3}; and
\item What is the computational complexity of the new method.
\end{itemize}

We show that the above estimation procedure has a highly nontrivial behavior. Statistical analysis remains nonetheless possible, but requires delicate results from discrete probability as well as novel statistical methods. We provide below theoretical justification of our approach for some basic cases.

\section{Applications to special cases}

Consider the case when $\varphi_2 \equiv 0$, and the spanning tree weight depends only on edge weights.\footnote{We alert the reader to the fact that analysis of just the expected weight of an ordinary (linear) MST for general graphs is a difficult problem~\citep{st02,frruth00}.}  As before, let $\mathcal{G} = \{ G \,|\, G \subseteq G_n \}$ be the chosen collection of (structured) subgraphs of $G_n$. Next, assume that for an arbitrary member $G = ( V , E ) \in \Gc$ the edge-cost set function $\varphi_1$, defined in (\ref{eq.1}), satisfies additionally
\begin{equation}
  \label{eq.4}
  \varphi_1 \bigr( \{ Y_e  \,|\, e \in E \} \bigr) \,=\, \varphi_1 \biggl( \sum\nolimits_{ e \in E } Y_e \biggr) \,,
\end{equation}
and that $\varphi_1$ is continuous and nondecreasing. This includes for example the important class of submodular functions that can be expressed as nondecreasing concave functions of sums (see e.g.,~\citep{stKr,goel}).

Based on the assumptions~\eqref{eq.1} and~\eqref{eq.4}, we can prove the following.
\begin{prop}
Let $F$ be a distribution function that is continuously differentiable at 0, having $F(0) = 0$ and $F'(0) > 0$. Suppose that $F$ has finite mean and variance. Assume that the cost function $\varphi$ satisfies~\eqref{eq.1} and~\eqref{eq.4} with $\varphi_2 \equiv 0$. Then for a minimum spanning tree of the complete graph $G_n$ it holds that

\begin{equation}
\lim_{n \rightarrow \infty} \mathbb{E}_F \, \varphi (\st (G_n)) \,=\, \varphi_1 ( \zeta (3) / F'(0) ) \,,
\end{equation}

\noindent where $\zeta$ is the Riemann Zeta function. Moreover, for any $\varepsilon > 0$,

\begin{equation}
\lim_{n \rightarrow \infty} \text{Pr}( | \varphi (\st (G_n)) - \zeta (3) / F'(0) | > \varepsilon ) \,=\, 0 \,.
\end{equation}
\end{prop}

The proof uses results from~\citep{fri85} and~\citep{st02}. Using this proposition, we will prove a consistency theorem for our estimator~(\ref{eq.3}) for the case of complete graphs\footnotemark[3] and a wide class of edge-dependent weight functions.

First, we need to introduce a special class of estimators.

\begin{defn}[\bf Boundary respecting estimators]
\label{def.bre}
As above, we assume that $F$ is a distribution function that is continuously differentiable at $0$, having $F(0) = 0$ and $F'(0) > 0$. Let $\mathcal{F}$ be some class of real-valued distribution functions that contains $F$. Suppose that we have a sequence of functions $\{ \Psi_{\mathcal{F}}^{(n)} \}_{n \geq 1}$ such that for each $n$ it holds that $\Psi_{\mathcal{F}}^{(n)} = ( \widehat{F}^{(n)} , \psi_{0}^{(n)} )$, where $\widehat{F}^{(n)}$ maps $\mathbb{R}^n \rightarrow \mathcal{F}$, and $\widehat{F}^{(n)}$ is differentiable at $0$ with the derivative $\psi_{0}^{(n)}$. Here $\psi_{0}^{(n)} ( X_1, X_2, \ldots , X_n )$ is a real-valued random variable itself.

Assume that there exists a real sequence $\{ r_n \}$ such that for any i.i.d.\ sample $X_1, X_2, \ldots , X_n$ generated from a distribution $F \in \mathcal{F}$, and any $\varepsilon > 0$, there exists a constant $C(\varepsilon, F) > 0$ such that for

\begin{equation}
\text{Pr} \bigr( \, | \psi_{0}^{(n)} (X_1, X_2, \ldots , X_n) - F'(0) | \,>\, \varepsilon \, \bigr) \,\leq\, \frac{\, C(\varepsilon, F) \,}{\, r_n \,} \,.
\end{equation}

If the sequence $r_n$ satisfies
\begin{equation}
\lim_{n \rightarrow \infty} r_n \,=\, \infty \,,
\end{equation}
we say that the estimator $\{ \Psi_{\mathcal{F}}^{(n)} \}_{n \geq 1}$ \emph{respects the boundary} of distribution $F$ from the class $\mathcal{F}$. In case the constant $C(F)$ above can be chosen independently of $F \in \mathcal{F}$, we say that the estimator $\{ \Psi_{\mathcal{F}}^{(n)} \}_{n \geq 1}$ \emph{respects the boundary uniformly} for distributions from the class $\mathcal{F}$.\hfill$\square$

\end{defn}

Such estimators actually exist---see~\citep{Balabdaoui2007} or~\citep{Alberts2003} for examples. It is necessary to remark here that a statistical question of constructing estimates that are consistent at boundary points can be tricky, and is certainly a \textbf{nonstandard} task. Blind use of standard methods can lead to incorrect results: many of the well-established estimation methods are consistent in integral norms such as $L_1$- or $L_2$-norms, or within the interior of the parameter spaces. Behavior of estimators at boundary points is substantially less studied, and the estimators that behave well at the boundary are usually not governed by conventional statistical results.

An an illustration, we note that in the setup of the Definition~\ref{def.bre}, the usual kernel density estimator gives a \emph{biased} estimate of $F'(0)$, even if one substantially restricts the space $\mathcal{F}$. Instead,~\citet{Alberts2003} proposes a modified kernel density estimator that has a correction for the bias on the boundary.

On the other hand, it is important to observe that Definition~\ref{def.bre} only requires that $F$ and $F^{'}$ are consistently estimated at the single boundary point 0; at other points $\widehat{F}^{(n)}$ may even be \emph{inconsistent}! This leaves a lot of opportunities for nonstandard constructions of estimators. Surprisingly enough, even inconsistent estimators are useful in our problem, as long as they respect the boundary.

\begin{theorem}[Consistency]
Let $G_n$ be a complete graph on $n$ vertices with random edge weights and let the cost function $\varphi$ satisfy~\eqref{eq.1} and~\eqref{eq.4} with $\varphi_2 \equiv 0$. Consider the problem of estimating the expected per vertex cost of an MST (using cost function $\varphi$) of $G_n$. Generate a complete auxiliary graph $G_{d(n)}^{'}$ on $d(n)$ vertices, via sampling the new edge weights $\{ Y_{e}^{'} \,|\, e \in E_{d(n)}^{'} \}$ from the distribution function $\widehat{F}^{( n (n-1)/2 )} ( \{ Y_{e} \,|\, e \in E_n \} )$. Suppose that $\{ \Psi_{\mathcal{F}}^{(n)} \}_{n \geq 1}$ respects the boundary for $F$.

1) Then, $\widehat{\Avcost} (G_n)$ is a consistent estimate, in the sense that for any $\varepsilon > 0$

\begin{equation}
\lim_{n \rightarrow \infty} \text{Pr} \biggr(\, \bigr| \, \widehat{\Avcost} (G_n) - \frac{\, 1 \,}{\, n \,} \varphi_1 \bigr(\st (G_n) \bigr) \,\bigr| \,>\, \varepsilon \,\biggr) \,=\, 0 \,.
\end{equation}

2) Much more than that, our auxiliary sample allows estimating the weight of the MST itself consistently in probability, i.e., for any $\varepsilon > 0$

\begin{equation}
\lim_{n \rightarrow \infty}  \text{Pr} \bigr( \, \bigr| \, d(n) \cdot \widehat{\Avcost} (G_n) -  \varphi_1 (\st (G_n)) \,\bigr| \,> \varepsilon \,\bigr) \,=\, 0 \,.
\end{equation}
\end{theorem}

The meaning of this theorem is that, for example, in the case of random complete graphs, one can consistently estimate some of their important characteristics by using just a small (but properly constructed) model of the initial large graph. In the particular case of spanning trees, one can have the number of vertices $d(n)$ grow to infinity \emph{arbitrarily slowly}, but still obtain asymptotically consistent estimates. This observation could be of much help in problems that require optimization on huge networks that would be practically intractable to treat as a whole.

\section{Related work and open problems}
In this section we first summarize some related work, and then discuss a list of open problems and challenges arising from this paper.

\subsection{Related work}
Random graph theory is a mature subject (see~\citep{belas}); but our interest is more specific. In particular, we draw upon work on estimating weights of (ordinary) MSTs dating back to~\citep{fri85}. For a good summary, and additional references we refer to the paper of~\citet{st02}.  \citet{bert90} studies a closely related but very different formulation, wherein he assumes that nodes may be present (or absent) with a certain probability. Based on this model, he studies what the expected weight of an MST might be. In contrast, we assume that the edge weights are random (according to specific law), and we study the expected value under a cost function strictly more general than the ordinary linear cost used for MSTs. Also note that in our framework one tends to build auxiliary graphs on $d(n) \ll n$ vertices, so our method is intended to works for graphs that are incomparably smaller than the original graph, while \citet{bert90} studies graphs on $O(n)$ nodes. 


To make our method practical, we depend on availability of an algorithm to solve the $\varphi$-MST problem on the auxiliary graph. For appropriate choices of the cost function $\varphi$, recent  algorithms such as those of~\citep{stKr} or \citep{jeg11,cvpr2010} might offer practical methods for tackling the subproblem on the auxiliary graphs. Additionally, there is a well-developed body on submodular optimization that we could tap into; see for instance \citep{chudak,fuji,iwata}. We note, however, that submodular set functions offer only one class of possible cost functions---if algorithms (or approximation algorithms) are available for other type of cost functions, we could benefit from those too---e.g., those in~\citep{murota}.

\subsection{Open Problems}
Since this is a position paper that also advances a new set of research problems, there are numerous aspects that remain to be studied. We highlight some of the important questions below.

An important open problem is to determine the types of deterministic or random graphs for which we can ensure consistency of the estimator from Theorem 1. There are fine probabilistic results on MSTs for several classes of random graphs, both asymptotic and finite sample (see~\citep{st02,frruth00} and references therein). Most notably, a lot is known about MSTs of cubes, and some other ``regular'' graphs. And, as some reflection shows, for such graphs, it is easy to check whether the estimator from Theorem 1 is consistent or not. But more generally, even if there is no hope to get closed form probabilistic results about the weight of the $\varphi$-MSTs, its proposed estimator may be expected to be consistent in many more interesting cases.

As shown in~\citep{st02}, the expected weight of a (linear) MST of an arbitrary connected graph $G$ can be represented as an integral of a function that depends on the Tutte polynomial of $G$. This observation leads us to conjecture that the expected weight of the $\varphi$-MST for submodular $\varphi$ might be representable as a Choquet integral involving Tutte polynomials. If this is the case, our estimators will also be randomized approximations of certain Choquet integrals, a curious byproduct.

As usual, it would be valuable to study rates of convergence of our estimators, as well as some basic properties such as asymptotic variance. The fact that these estimators can be consistent even when they are based on a ``small'' graph (with $d(n) \ll n$ vertices), is promising since it provides theoretical grounds for replacing processing on giant networks by processing suitably constructed, smaller networks. Results on variance and rates of convergence of the estimators will contribute towards judging actual accuracy of such replacements.

Since we expect our estimation to work on large graphs, it is crucial that we be able to minimize the cost function $\varphi$ efficiently, at least on the auxiliary graph $G_{d(n)}^{'}$. This raises the cornerstone question: for which types of cost functions $\varphi$ (submodular, monotone, etc.) does there exist an efficient optimization method for finding (at least approximately) the desired minimum cost structure (spanning tree, path, etc.) that simultaneously also respects our statistical estimation procedure. The present short paper suggests that this class of cost functions is rich (at least infinite-dimensional).

Finally, we close by mentioning that even though we illustrated only spanning trees, the same argument extends to obtaining estimators for any other graphical structures such as paths, cuts, etc., as long as suitable estimators are available for corresponding linear cost functions. More challengingly, we wish to consider deriving conditions on $\varphi_1$ and $\varphi_2$ in the decomposition~\eqref{eq.1}, under which one obtains consistent estimators.

\bibliographystyle{abbrvnat}
\bibliography{graphs}

\begin{thebibliography}{15}
\providecommand{\natexlab}[1]{#1}
\providecommand{\url}[1]{\texttt{#1}}
\expandafter\ifx\csname urlstyle\endcsname\relax
  \providecommand{\doi}[1]{doi: #1}\else
  \providecommand{\doi}{doi: \begingroup \urlstyle{rm}\Url}\fi

\bibitem[Alberts and Karunamuni(2003)]{Alberts2003}
T.~Alberts and R.~J. Karunamuni.
\newblock A semiparametric method of boundary correction for kernel density
  estimation.
\newblock \emph{Statistics and Probability Letters}, 61\penalty0 (3):\penalty0
  287--298, 2003.

\bibitem[Balabdaoui(2007)]{Balabdaoui2007}
F.~Balabdaoui.
\newblock Consistent estimation of a convex density at the origin.
\newblock \emph{Mathematical Methods of Statistics}, 16:\penalty0 77--95, 2007.
\newblock ISSN 1066-5307.

\bibitem[Bertsimas(1990)]{bert90}
D.~J. Bertsimas.
\newblock The probabilistic minimum spanning tree problem.
\newblock \emph{Networks}, 20\penalty0 (3):\penalty0 245--275, 1990.

\bibitem[Bollob\'as(2001)]{belas}
B.~Bollob\'as.
\newblock \emph{{Random Graphs}}.
\newblock Cambridge University Press, 2001.

\bibitem[Chudak and Nagano(2007)]{chudak}
F.~A. Chudak and K.~Nagano.
\newblock Efficient solutions to relaxations of combinatorial problems with
  submodular penalties via the {L}ov\'asz extension and nonsmooth convex
  optimization.
\newblock In \emph{SODA}, 2007.

\bibitem[Frieze(1985)]{fri85}
A.~M. Frieze.
\newblock On the value of a random minimum spanning tree problem.
\newblock \emph{Discrete Applied Mathematics}, 10:\penalty0 47--56, 1985.

\bibitem[Frieze et~al.(2000)Frieze, Ruszink\'o, and Thoma]{frruth00}
A.~M. Frieze, M.~Ruszink\'o, and L.~Thoma.
\newblock A note on random minimum length spanning trees.
\newblock \emph{Electronic Journal of Combinatorics}, 2000.

\bibitem[Fujishige(2005)]{fuji}
S.~Fujishige.
\newblock \emph{Submodular functions and optimization}, volume~58 of
  \emph{Annals of Discrete Mathematics}.
\newblock Elsevier Science, 2005.

\bibitem[Goel et~al.(2010)Goel, Tripathi, and Wang]{goel}
G.~Goel, P.~Tripathi, and L.~Wang.
\newblock {Optimal Approximation Algorithms for Multi-agent Combinatorial
  Problems with Discounted Price Functions}.
\newblock In \emph{Foundations of Software Technology and Theoretical Computer
  Science}, 2010.

\bibitem[Iwata and Nagano(2009)]{iwata}
S.~Iwata and K.~Nagano.
\newblock Submodular function minimization under covering constraints.
\newblock In \emph{FOCS}, 2009.

\bibitem[Jegelka and Bilmes(2011{\natexlab{a}})]{cvpr2010}
S.~Jegelka and J.~A. Bilmes.
\newblock Submodularity beyond submodular energies: coupling edges in graph
  cuts.
\newblock In \emph{Computer Vision and Pattern Recognition (CVPR)}, June
  2011{\natexlab{a}}.

\bibitem[Jegelka and Bilmes(2011{\natexlab{b}})]{jeg11}
S.~Jegelka and J.~A. Bilmes.
\newblock Approximation bounds for inference using cooperative cuts.
\newblock In \emph{International Conference on Machine Learning (ICML)},
  2011{\natexlab{b}}.

\bibitem[Murota(2003)]{murota}
K.~Murota.
\newblock \emph{Discrete Convex Analysis}.
\newblock SIAM, 2003.

\bibitem[Steele(2002)]{st02}
J.~M. Steele.
\newblock Minimum spanning trees for graphs with random edge lengths.
\newblock In \emph{In Mathematics and Computer Science II: Algorithms, Trees,
  Combinatorics and Probabilities, Birkh\"auser}, pages 223--245, 2002.

\bibitem[Stobbe and Krause(2010)]{stKr}
P.~Stobbe and A.~Krause.
\newblock Efficient minimization of decomposable submodular functions.
\newblock In \emph{NIPS}, 2010.

\end{thebibliography}

\end{document}